%% file: main.tex
\documentclass[11pt,letterpaper]{article}
\usepackage[margin=1in]{geometry}
\usepackage{lmodern}
\usepackage[T1]{fontenc}
\usepackage[utf8]{inputenc}

\input{math_commands.tex}

\usepackage[colorlinks,citecolor=blue,bookmarks=true,linktocpage]{hyperref}
\usepackage{url}
\usepackage[svgnames]{xcolor}
\definecolor{cnavy}{rgb}{0,0,.5}
\definecolor{cmaroon}{rgb}{0.7,0.1,0.2}

\hypersetup{
    colorlinks = True,
    linkcolor=cmaroon,
    citecolor=cnavy,
    urlcolor=magenta
}

\usepackage[style=alphabetic, maxbibnames=15, maxcitenames=3, natbib=true,
  maxalphanames=6, backend=biber, sorting=nty, backref=true]{biblatex}
\DefineBibliographyStrings{english}{%
  backrefpage = {page},%
  backrefpages = {pages},%
}
\usepackage{csvsimple}
\usepackage{booktabs}
\usepackage{subcaption}
\usepackage{nicefrac}
\usepackage{ifthen}
\usepackage[capitalise]{cleveref}
\crefname{figure}{Figure}{Figures}
\usepackage[normalem]{ulem}
\usepackage{graphics}
\usepackage{multirow}

\newcommand{\0}{\boldsymbol{0}}

\newcommand{\y}{\mathbf{y}}
\newcommand{\hy}{\widehat{\mathbf{y}}}
\newcommand{\sample}[2]{\y_{#1 : #2}}
\newcommand{\hsample}[2]{\hy_{#1 : #2}}
\newcommand{\icsample}[3]{\sample{#1}{#2}^{(#3)}}

\newcommand{\inputResidualLayer}[1]{\mathsf{InputResidualLayer}(#1)}
\newcommand{\stackedTransformer}[1]{\mathsf{StackedTransformer}(#1)}
\newcommand{\outputResidualLayer}[1]{\mathsf{OutputResidualLayer}(#1)}

\newcommand{\ctx}{\sample{1}{L}}
\newcommand{\addctx}[2][]{\ifthenelse{\equal{#1}{}}{\icsample{1}{T_{#2}}{#2}}{\y_{#1}^{(#2)}}}
\newcommand{\separator}{\boldsymbol{\sigma}}

\newcommand{\patched}[2]{\y_{p(#1 - 1) + 1 : p #1}^{(#2)}}
\newcommand{\patch}[2]{\tilde{\y}_{#1}^{(#2)}}

\newcommand{\pmask}[2]{\tilde{\mathbf{m}}_{#1}^{(#2)}}
\newcommand{\pmaskdot}[2]{\dot{m}_{#1}^{(#2)}}

\newcommand{\token}[2]{\mathbf{t}_{#1}^{(#2)}}
\newcommand{\mtoken}[2][]{\ifthenelse{\equal{#1}{}}{\tilde{\mathbf{t}}^{(#2)}}{\tilde{\mathbf{t}}_{#1}^{(#2)}}}

\newcommand{\stout}[2]{\mathbf{o}_{#1}^{(#2)}}

\newcommand{\horizon}{\hsample{L+1}{L+H}}
\newcommand{\phorizon}[2]{\hy_{p #1 + 1: p #1 + h}^{(#2)}}
\newcommand{\target}{\sample{L+1}{L+H}}

\input{defs.tex}

\newcommand*\samethanks[1][\value{footnote}]{\footnotemark[#1]}
\newcommand\extrafootertext[1]{%
    \bgroup
    \renewcommand\thefootnote{\fnsymbol{footnote}}%
    \renewcommand\thempfootnote{\fnsymbol{mpfootnote}}%
    \footnotetext[0]{#1}%
    \egroup
}

\author{Abhimanyu Das\thanks{Google Research.
    }
\and
Matthew Faw\thanks{The University of Texas at Austin. Correspondence to: \textsf{\{\href{mailto:matthewfaw@utexas.edu}{matthewfaw}@utexas.edu\}}. This work was done while the author was a Student Researcher at Google Research.}
\and
Rajat Sen\samethanks[1]
\and
Yichen Zhou\samethanks[1]}

\title{In-Context Fine-Tuning for Time-Series Foundation Models}

\begin{document}

\maketitle

\begin{abstract}
\extrafootertext{Authors listed in alphabetical order.}
Motivated by the recent success of time-series foundation models for zero-shot forecasting, we present a methodology for \emph{in-context fine-tuning} of a time-series foundation model. In particular, we design a pretrained foundation model that can be prompted (at inference time) with multiple time-series examples, in order to forecast a target time-series into the future. Our foundation model is specifically trained to utilize examples from multiple related time-series in its context window (in addition to the history of the target time-series) to help it adapt to the specific distribution of the target domain at inference time.  We show that such a foundation model that uses in-context examples at inference time can obtain much better performance on popular forecasting benchmarks compared to supervised deep learning methods, statistical models, as well as other time-series foundation models.  Interestingly, our in-context fine-tuning approach even rivals the performance of a foundation model that is explicitly fine-tuned on the target domain.
\end{abstract}

\section{Introduction}
Time-series data is ubiquitous in several domains such as retail, finance, manufacturing, healthcare, and natural sciences. In many of these domains, time-series forecasting, i.e., predicting time-series into the future, is a critical problem -- for example, in applications like retail forecasting, climate and weather predictions, and traffic forecasting. In the last decade, deep learning approaches~\citep{salinas2020deepar, oreshkin2019n, sen2019think} have become popular in forecasting, often outperforming statistical approaches like ARIMA~\citep{box1968some}.
However, until recently, deep learning approaches for forecasting have adhered to the traditional supervised machine learning framework of having to train a forecasting model on task-specific training data, before being able to perform forecasting for that task. 
On the other hand, in Natural Language Processing (NLP), Large Language Models (LLMs)~\citep{radford2019language, brown2020language} have shown the promise of foundation models: a single pretrained model can perform well and adapt to tasks like translation, code generation, text summarization during inference time in a zero-shot or few-shot manner. 

Motivated by the success in NLP, 
there has been significant work in recent years on time-series foundation models for forecasting, ranging from re-purposing LLMs directly for forecasting~\citep{gruver2023large} to fine-tuning pretrained LLMs on time-series data~\citep{zhou2023one, chang2023llm4ts} to pretraining time-series foundation models from scratch~\citep{das2023decoder,goswami2024moment,woo2024unified,ansari2024chronos,garza2023timegpt}. The last approach in particular has been shown to obtain strong zero-shot accuracy, rivaling the best supervised models trained specifically for the target datasets. 

Several of these papers~\citep{zhou2023one,ansari2024chronos,goswami2024moment} have shown an opportunity for further accuracy improvement by fine-tuning of their pretrained models on target datasets. However, this breaks the zero-shot paradigm that precisely makes these time-series foundation models so appealing to practitioners who do not want to build training pipelines. This raises a natural question: \emph{Can we recover the benefits of fine-tuning a time-series foundation-model by providing examples from a target dataset at inference time?}

At the same time, the first iterations of these foundation models lack some of the desirable features of LLMs with respect to \textit{in-context learning}: the zero-shot performance of an LLM can be greatly improved \textit{at inference time} by using its context window for prompting techniques such as few-shot~\citep{brown2020language}, chain-of-thought~\citep{wei2022chain} or instruction tuning~\citep{wei2021finetuned}. These papers have shown emergent in-context learning abilities for LLMs. In particular, if we prompt them with related examples, demonstrations and instructions, then ask a specialized question, the model is able to reason similarly for the question at hand. 

In this work, we study a methodology to enable similar in-context ability for a time-series foundation model in terms of being able to prompt the model with time-series examples from the target domain, and recover the benefits of domain-specific fine-tuning. We refer to this as \emph{in-context fine-tuning}.\footnote{Terminology: In the LLM domain, this notion is also called ``few-shot learning''~\citep{brown2020language}, ``few-shot prompting''~\citep{ye2022unreliability}, or ``in-context tuning''~\citep{chen2021meta}. Also, borrowing from LLM literature, we will refer to the generic ability of pretrained foundation models to learn from information in their context-window at inference time as ``in-context learning''. Additionally, we will refer to pretrained models that do not need gradient-updates via explicit training or tuning for an unseen target dataset as ``zero-shot''.}

We train a foundation model that lets us forecast a time-series by providing in its context window not just the historical values of the time-series, but also examples from other related time-series that could help the model adapt, \textit{at inference time}, to the distribution of the target time-series.
For example, consider a highway traffic prediction system that stores hourly data from the last week, in order to predict the future hourly traffic for a particular highway. Consider a time-series foundation model that has not seen data in pretraining that captures the temporal patterns in this traffic data. Then, simply prompting the model with the previous week's traffic time-series for that highway might not be enough to obtain accurate zero-shot performance. However, adding to the prompt historical traffic data from other highways and weeks, might help the model better adapt to the traffic data distribution, and improve the target accuracy significantly.

\begin{figure}
    \centering
    \begin{subfigure}[t]{0.45\textwidth}
        \centering
        \includegraphics[width=\textwidth]{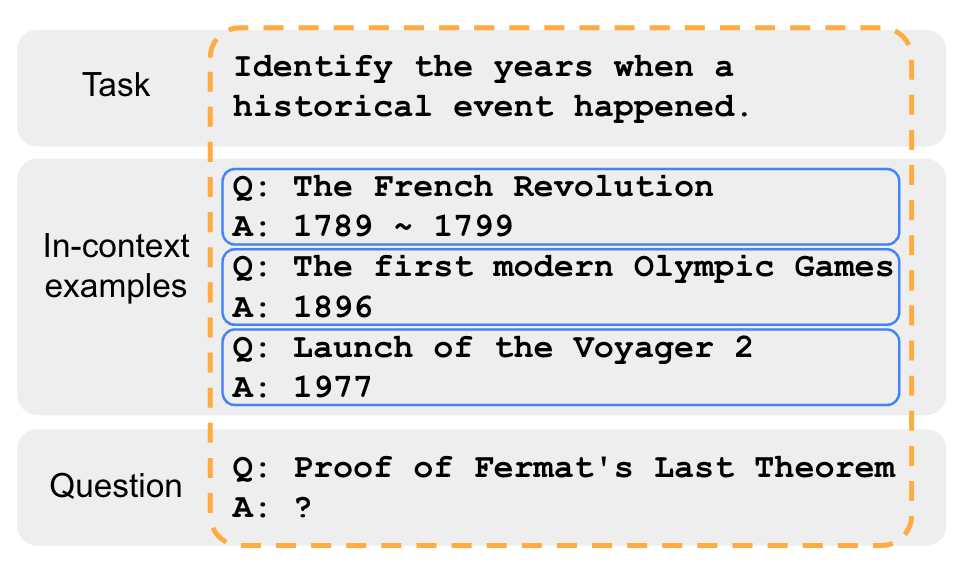}
    \end{subfigure}%
    ~ 
    \begin{subfigure}[t]{0.45\textwidth}
        \centering
        \includegraphics[width=\textwidth]{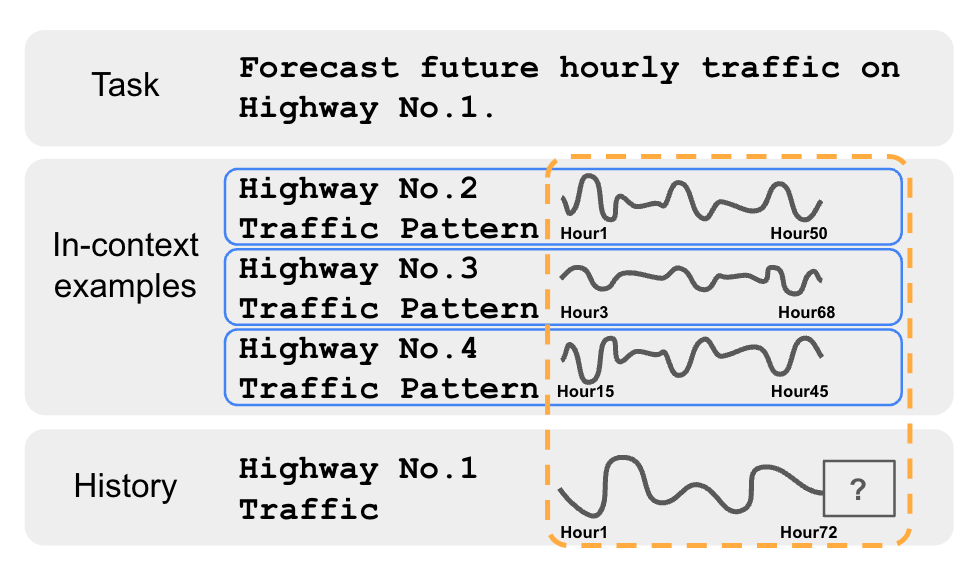}
    \end{subfigure}
    \caption{Analogous to few-shot prompting of a foundation LLM (left), we train a time-series foundation model to support few-shot prompting with an arbitrary number of related in-context time-series examples (right). The dashed box encloses the full context window/prompt.}    
    \label{fig:motivatingExample}
\end{figure}

To summarize, the main contributions of our paper are as follows:

{\bf (i)~} We introduce the study of in-context fine-tuning for time-series foundation models, and propose the use of prompts that not only include the usual history of the target time-series for forecasting, but also include related time-series examples in-context.

{\bf (ii)~} We pretrain a time-series foundation model to be able to effectively utilize these in-context time-series examples mentioned above. Our training is decoder-only~\citep{liu2018generating} and can adapt not only to any context, horizon pair (up to a certain maximum context) but also to any number of supplementary time-series examples (again up to a certain maximum number of examples). Appropriately trained models can then learn to borrow patterns from these related examples to do better on the original forecasting task.

{\bf (iii)~} We empirically evaluate the benefits of in-context fine-tuning using our foundation model. Using evaluations on popular forecasting benchmarks, we show that in-context fine-tuning can lead to better zero-shot performance on popular forecasting benchmarks as compared to supervised deep learning methods, statistical models as well as other foundation models. In particular, it obtains up to 25\% better performance than a state-of-the-art time-series foundation model and other supervised deep learning and statistical baselines. Surprisingly, it even slightly improves upon the performance of a time-series foundation model that is specifically fine-tuned to the target datasets.

\section{Related Work}

As mentioned previously, there has been a spurt of recent work on time-series foundation models for forecasting. These approaches can be broadly divided into three categories. (i) Prompting LLMs like GPT-4 to directly predict the future of a numerical series encoded as text. This was investigated in LLMTime~\citep{gruver2023large}; despite the initial promise subsequent works have shown that such approaches can be lacking in accuracy~\citep{woo2024unified, das2023decoder}. (ii) Fine-tuning pretrained LLMs like GPT2 on time-series data with adapter layers~\citep{zhou2023one, chang2023llm4ts}. These approaches have mostly been shown to work well in the dataset-to-dataset transfer learning setting (rather than in the zero-shot setting), and they are also disadvantaged from having to use excessively large models due to their LLM backbones. (iii) Pretraining transformer based models from scratch on huge volumes of time-series data, which seems to be the most promising approach towards times-series foundation models~\citep{das2023decoder, garza2023timegpt, ansari2024chronos, woo2024unified, goswami2024moment}. Indeed, some of these models have shown superior zero-shot accuracy when compared to supervised deep forecasters and statistical methods even on datasets that are outside of their pretraining set.

Some of the above papers, e.g.,~\citep{ansari2024chronos,goswami2024moment}, have additionally shown that their pretrained models' performance can be further improved by fine-tuning the model on examples from the target dataset. 
While this supervised fine-tuning results in improved per-task accuracy, this practice breaks the zero-shot paradigm in terms of requiring extra training on the target dataset.

In the NLP domain, a defining property of a foundation LLM is its ability to be further adapted to domain-specific tasks through either fine-tuning or prompting. In particular, LLMs have been shown to perform \emph{in-context learning} on a variety of downstream NLP tasks by prompting them with a natural language instruction~\citep{radford2019language} and a few demonstrations or examples of the task. This phenomenon is also referred to as \emph{few-shot learning}~\citep{brown2020language}. Subsequent works~\citep{min2021metaicl,chen2021meta} have proposed fine-tuning a pretrained LLM to obtain better performance on few-shot learning prompts. Other papers~\citep{min2022rethinking,wei2023larger} have empirically investigated how few-shot learning works in LLMs. More recently, \citet{shi2023context} explored a similar idea for in-context pretraining, where they pretrain an LLM on sequences of related documents. This in-context learning ability is widely recognized as being associated with the stacked transformers used in the LLMs, and their theoretical properties are studied in a more precise sense~\citep{garg2022can,von2023transformers,ahn2024transformers} for simpler function classes such as linear regression.

Despite the commonality between time-series foundation models and LLMs, it is not obvious how (or even if) the phenomenon of few-shot learning for NLP tasks carry over to the time-series setting. There is no clear definition of few-shot learning in terms of a time-series foundation model. In fact, prior pretrained time-series foundation models like~\citep{ansari2024chronos, das2023decoder, garza2023timegpt} do not provide a clear opportunity to be prompted with anything apart from the past values of a time-series in the context window.

\section{Problem Definition}

\begin{figure}[t]
    \centering
    \includegraphics[width=.4\textwidth]{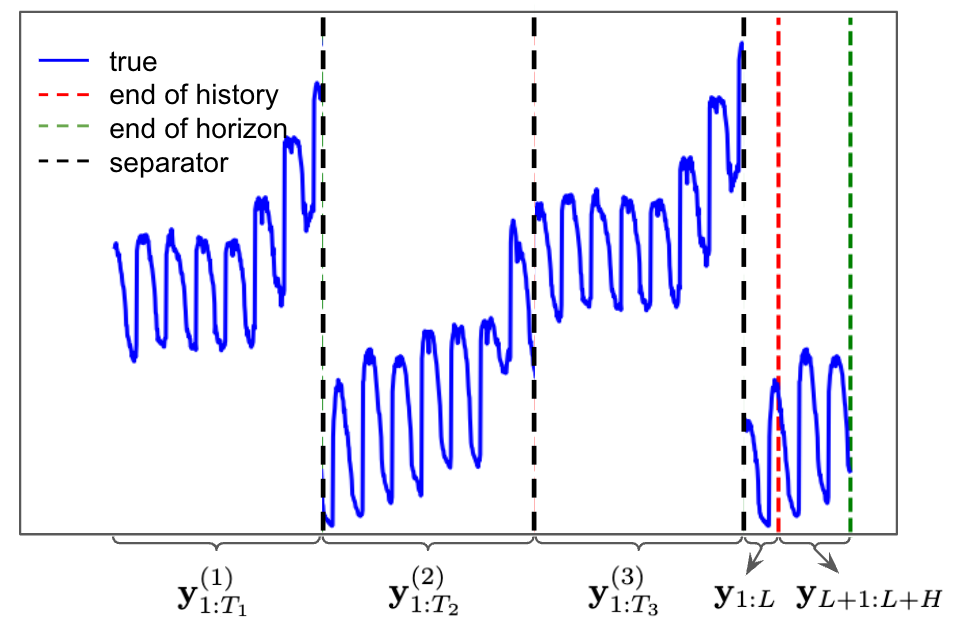}
    \caption{An example prediction task $(\{\addctx{1},\addctx{2}, \addctx{3}, \ctx\}, \target)$. The three black dashed lines (separators) separate the three in-context examples $\{\addctx{i}\}_{i\in[3]}$ and the history $\ctx$. The goal is to predict the horizon $\target$ of the history $\ctx$.}
    \label{fig:chunkedExample}
\end{figure}
\begin{figure}[ht]
    \centering
    \begin{subfigure}[t]{0.30\textwidth}
        \centering
        \includegraphics[width=\textwidth]{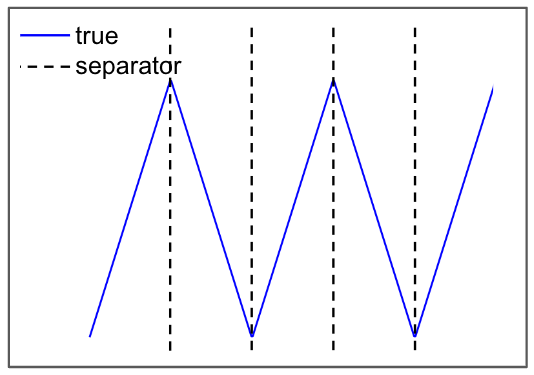}
        \caption{Multiple linear trends.}
        \label{subfig:linearTrendsVS}
    \end{subfigure}\hspace{.1\textwidth}
    \begin{subfigure}[t]{0.3\textwidth}
        \centering
        \includegraphics[width=\textwidth]{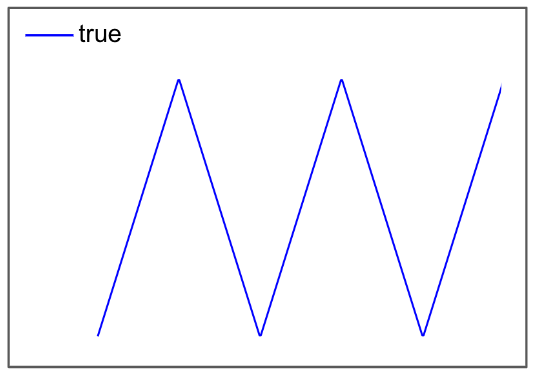}
        \caption{A triangular wave.}
        \label{subfig:triangleWaveVS}
    \end{subfigure}
    \caption{A prediction task with two forms of concatenation: in \cref{subfig:linearTrendsVS}, we concatenate with separators, and in \cref{subfig:triangleWaveVS}, we concatenate without separators. Concatenating in-context examples together without separators can confuse the model: multiple linear trends look like a triangular wave if concatenated na\"ively.}
    \label{fig:triangleWaveVSLinearTrends}
\end{figure}

Time-series foundation models aim to build a general purpose forecaster that can take in a past \textit{history} of a target forecasting task, $\ctx = \{y_1, y_2, \cdots y_L\}$, where we look back $L$ time-steps and map them to a forecast $\horizon$, for a horizon length of $H$. The aim is to have $\horizon$ as close as possible to the unseen future $\target$ according to some well defined error metric. Such a model can be thought of as a function,
\begin{equation}
    g: \ctx \rightarrow \horizon \label{eq:usual_tfm}
\end{equation}
which is capable for handling different values of $L$ and $H$.

In this work, we aim to further enhance the abilities of such models by enriching their context. In addition to the target task's history $\ctx$, we add up to $n-1$ \textit{in-context examples} of the form $\{\addctx{1}, \addctx{2}, \cdots \addctx{n-1}\}$ that can represent the past time-points of other related time-series (with possibly varying lengths $T_1,\cdots, T_{n-1}$). In the case of our motivating example of highway traffic prediction, $\ctx$ is a week of hourly traffic data on that highway, and $\{\*y^{(1)}_{1:T_1}, \*y^{(2)}_{1:T_2}, \cdots \*y^{(n-1)}_{1:T_{n-1}}\}$ are traffic data on $n-1$ nearby highways. We plot an example prediction task with three in-context examples in \cref{fig:chunkedExample}.

Therefore, the enhanced forecasting problem is aimed at training a model $f$,

\begin{equation}
    f: \left(\*y^{(1)}_{1:T_1}, \*y^{(2)}_{1:T_2}, \cdots \*y^{(n-1)}_{1:T_{n-1}}, \ctx\right) \rightarrow \horizon \label{eq:new_tfm}.
\end{equation}

As before, our time-series foundation model should be able to handle different values of $L$ and $H$. Additionally it should be able to support any number of in-context examples ($n-1$) ranging from zero to a maximum value. With some abuse of notation, let us index the target task's forecasting history and horizon as the $n$-th example i.e. $\*y^{(n)}_{1:T_n} := \*y_{1:L+H},$ where $T_n = L + H$. Therefore, our decoder-only model will work with $n$ examples of the form $\{\*y^{(1)}_{1:T_1}, \*y^{(2)}_{1:T_2}, \cdots,\*y^{(n)}_{1:T_{n}}\}$ which are drawn from related time-series. Henceforth, we will refer to $\{\*y^{(i)}_{1:T_i}\}_{i=1}^n$ as the \textit{context} (synonymous with prompt) supplied to the model. 

\section{Model Architecture}

\begin{figure}[ht]
    \centering
    \includegraphics[width=.75\textwidth]{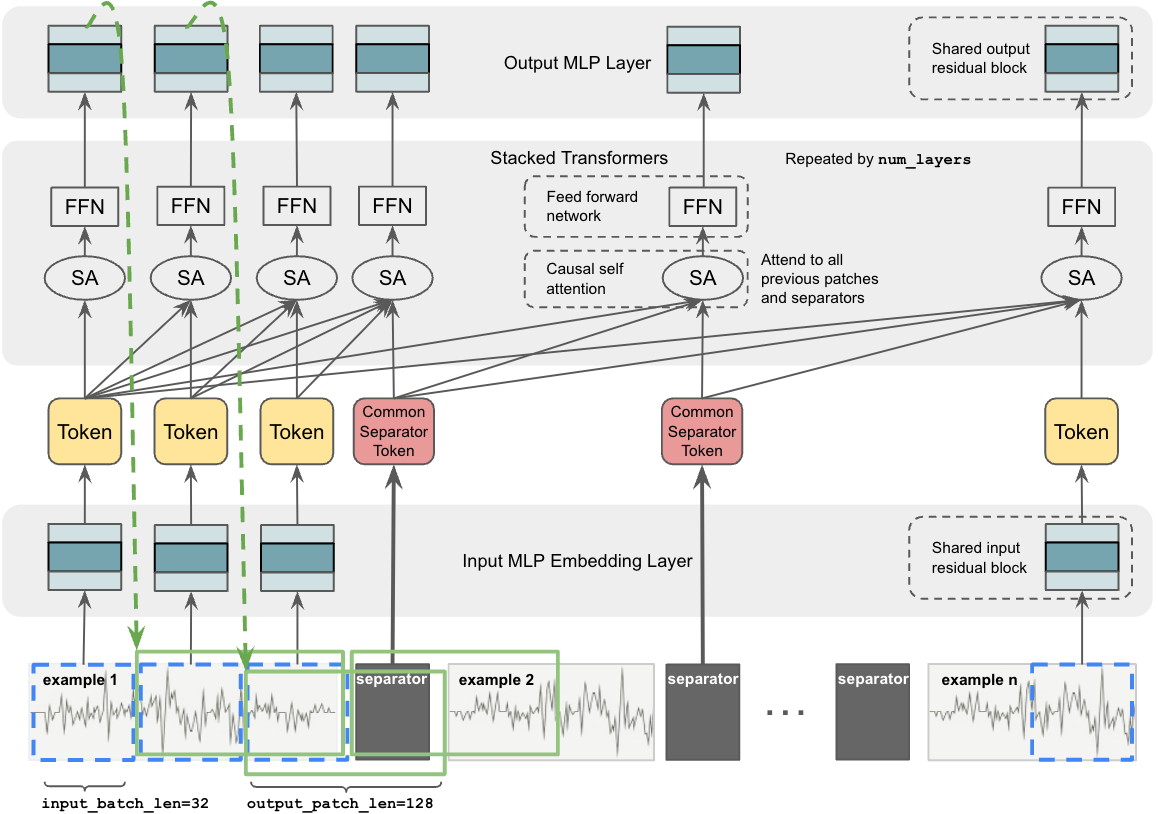}
    \caption{Our decoder-only architecture for time-series prediction with in-context examples.}
    \label{fig:architecture}
\end{figure}

Motivated by the strong zero-shot performance achieved by stacked transformer models in decoder-only mode for time-series forecasting, we propose to adapt a base TimesFM model~\citep{das2023decoder} to leverage the additional information available via in-context examples. In particular, we pretrain TimesFM in its original fashion to obtain a base checkpoint. We then modify the model architecture and continue pretraining from the base checkpoint using training data with in-context examples (we call this phase \textit{continued pretraining}) to obtain a new pretrained foundation model \textit{\ours}. The base TimesFM checkpoint that we start from will be referred to as \textit{\oursbase}.

Adapting their model architecture to make use of the in-context examples is somewhat delicate, and requires modifications to the original model. A depiction of our proposed model architecture is given in \cref{fig:architecture}. As in their model, our model partitions each example into non-overlapping input \emph{patches}, and uses a shared input residual block (a one-hidden layer perceptron with skip connection, see~\citet{das2023longterm}), to embed each patch as a token before feeding the tokens into the stacked transformers in a decoder-only fashion. The output embeddings are mapped to the next output patches via another shared output residual block.

To teach the model to use the new in-context examples, we adapt the original TimesFM architecture to better handle (1) the in-context example separators, (2) the cross-example attention, and (3) the positional encoding. Despite these changes, we are still able to leverage the TimesFM (base) checkpoint, which was pretrained for forecasting given just the history of the target time-series. We describe the key details of our architecture design below.

\subsection{Separators for In-context examples}
Our context window contains in-context examples from different time-series. Hence the model needs to be able to separate these, since
na\"ive concatenation can confuse the model. Consider the example in \cref{fig:triangleWaveVSLinearTrends}. If we na\"ively concatenate multiple in-context  examples (e.g., linear trends, \cref{subfig:linearTrendsVS}) together, then the combination of these trends may appear to the model as an entirely different time-series (e.g., a triangle wave, \cref{subfig:triangleWaveVS}). Therefore, we choose to insert a common learnable separator token after each in-context example. We visually depict these separators as the dashed lines in \cref{subfig:linearTrendsVS}. When feeding examples to the decoder, we sequentially pass each tokenized patch of each time-series example to the model, followed by the separator token at the end of an example.
This process is depicted in \cref{fig:architecture}.

\subsection{Cross-example Attention} 
In order to allow our model to distinguish between different in-context examples, we allow the transformer to attend (causally) to all previous patches including the separator tokens. 
Note that, if the model did not attend to the separator tokens, then we could never hope to distinguish between the two scenarios from \cref{subfig:linearTrendsVS} and \cref{subfig:triangleWaveVS}. By attending to the previous separator tokens, the model can potentially distinguish how many in-context examples have been processed so far. 

Although at the input to the stacked transformer we use a common separator token to separate the examples, the output tokens corresponding to the positions of these separator tokens can play a much more nuanced role as we proceed through the subsequent transformer layers. As the output tokens corresponding to these separator tokens causally attend to all previous tokens, after several transformer layers these tokens can, for instance, potentially summarize the information in all the patches corresponding to their example and/or convey the separation boundaries of the different in-context examples to the model.

\subsection{Positional Encoding}
Based on the findings in~\citet{haviv2022transformer}, we create the pretrained TimesFM (base) checkpoint with No Positional Encodings (NoPE), in contrast to the absolute positional encodings~\citep{vaswani2017attention} used in the original TimesFM model. We note that we can achieve the same accuracy reported in the original TimesFM paper without using any positional encodings. Indeed it has been hypothesized in~\citet{haviv2022transformer} that the presence of causal attention itself can encode positional information when there are more than one stacked transformer layers.

The advantages of NoPE for our continued pretraining are two fold: (i) NoPE models usually have better length generalization, which is particularly important here since we increase the prompt length by adding in-context examples to the context. (ii) If we use the original absolute positional encodings used in~\citep{das2023decoder}, the meaning of these positional encodings in the base model would be different from their meaning during the continued pretraining with in-context examples. This could cause problems for the continued pretraining phase.

\subsection{Overall Model}

Since our model builds upon the TimesFM architecture \citep{das2023decoder}, we introduce a similar notation style for ease of exposition.
The model processes in-context examples in the following fashion. Starting with an example input $\{\addctx{1},\ldots,\addctx{n}\}$, each example $\addctx{i}$ is partitioned into input patches of length $p$:
\begin{align*}
    \patch{j}{i} = \patched{j}{i} \quad \forall j \in [\lceil \nicefrac{T_i}{p} \rceil] \text{ and } i \in [n].
\end{align*}
As in \citep{das2023decoder}, our model takes an additional padding mask $\mathbf{m}_{1 : T_i}^{(i)}$ to ensure that it makes good predictions on time-series which are not a multiple of the patch length $p$. Analogously to the partitioning of the example inputs, we partition the padding masks as:
\begin{align*}
    \pmask{j}{i} = \mathbf{m}_{p(j-1)+1:pj}^{(i)} \quad \forall j \in [\lceil \nicefrac{T_i}{p} \rceil] \text{ and } i \in [n].
\end{align*}
Given these patches and masks, we feed each patch $\patch{j}{i}$ through a common MLP embedding layer to obtain tokens:
\begin{align*}
    \token{j}{i} = \inputResidualLayer{\patch{j}{i}\odot(1-\pmask{j}{i})} \quad \forall j \in [\lceil \nicefrac{T_i}{p} \rceil] \text{ and } i \in [n].
\end{align*}
We will slightly abuse notation by denoting the separator token $\separator$ as
$\token{\ceil{\nicefrac{T_i}{p}} + 1}{i} = \separator,$
and let the mask for the separator token $\pmask{\ceil{\nicefrac{T_i}{p}} + 1}{i} = \0$ (i.e., the separator tokens are never masked).
After tokenizing the input patches, we feed the tokens, together with a learnable separator token $\separator$, autoregressively to the stacked transformer layers in decoder-only mode. We take $\pmaskdot{j}{i}$ to be the last entry of $\pmask{j}{i}$\footnote{Intuitively, $\pmaskdot{j}{i}$ indicates whether or not patch $\patch{j}{i}$ is masked from the right. We attend only to patches which are not padded from the right, and have at least one unpadded values (see Appendix~\ref{app:model})}, and denote the sequence of token/mask pairs corresponding to example $i$ as
\begin{align*}
    \mtoken[1:j]{i} 
    = ((\token{1}{i}, \pmaskdot{1}{i}), \ldots, (\token{j}{i}, \pmaskdot{j}{i}))
    \quad \forall j \in [\lceil \nicefrac{T_i}{p} \rceil + 1] \text{ and } i \in [n].
\end{align*}
Then, the output of the stacked transformer layer for token $\token{j}{i}$ can be described as:
\begin{align*}
    \stout{j}{i} = \stackedTransformer{\mtoken[1:\ceil{\nicefrac{T_i}{p}}+1]{1}, \ldots, \mtoken[1:\ceil{\nicefrac{T_i}{p}}+1]{i-1}, \mtoken[1:j]{i}}
    \quad \quad \forall j \in [\lceil \nicefrac{T_i}{p} \rceil+1] \text{ and } i \in [n].
\end{align*}
We emphasize the output $\stout{j}{i}$ for token $\token{j}{i}$ defined above depends on (i) all previous (unmasked) tokens $\token{j'}{i'}$, $i'<i$ and $j'\in [\ceil{\nicefrac{T_{i'}}{p}}]$, (ii) the $i-1$ separator tokens $\token{\ceil{\nicefrac{T_{i'}}{p}}+1}{i'} = \separator$ for $i'<i$, and (iii) the tokens $\mtoken[1:j]{i}$ for the current example.

Finally, we feed the outputs $\stout{j}{i}$ from the stacked transformer through a residual block to obtain the predicted time-series:
\begin{align*}
    \phorizon{j}{i} = \outputResidualLayer{\stout{j}{i}}
    \qquad \forall j \in [\ceil{\nicefrac{T_i}{p}}] \text{ and } i \in [n].
\end{align*}
This corresponds to the model's prediction of the next $h$ steps (output patch length) of $\addctx[pj + 1 : pj+h]{i}$.

\subsection{Loss Function}

Similar to ~\citep{das2023decoder}, we use Mean Squared Error (MSE) as our point forecast loss:
\begin{align*}
\mathsf{TrainLossPerContext} = \frac{1}{\sum_{i=1}^n\lceil \nicefrac{T_i}{p} \rceil}\sum_{i = 1}^n \sum_{j=1}^{\lceil \nicefrac{T_i}{p} \rceil} \| \phorizon{j}{i}  - \mathbf{y}_{pj + 1: p j + h}^{(i)}\|^2.
\end{align*}

\section{Pretraining Data}
\label{sec:training}

As mentioned before, we start with 
\oursbase~which was pretrained on a diverse corpus of about 400B time-points. Please see Table~\ref{tab:train_data} in Appendix~\ref{app:model} and ~\citet{das2023decoder} for more details on the datasets. We then continue pretraining it on training data containing in-context examples.

{\bf Context Generation.} We convert individual datasets to generate \textit{contexts} with in-context examples that the model sees during the continued pretraining. Recall that the original TimesFM model is trained up to a maximum history length of $L_{max}=512$. During the training of \oursbase~a time-series of length $T = L_{max} + h$ is loaded for back propagation where $h=128$ is the output patch length. Therefore, we choose $T$ as the maximum length of our $n$ in-context examples. For any time-series in a particular dataset, we use windowing with a shift of $1$ to generate examples of length $T$ i.e. for a time-series $\*y_{1:M}$ the possibles examples are $\left\{\*y_{1:T}, \*y_{2:T+1}, \cdots \*y_{M-T+1:M}\right\}$. For time-series that are less than $T$ in length, we generate padded examples as detailed in Appendix~\ref{app:model}. Now these examples are packed in groups of $n$ to form the context. We consider two kinds of grouping:

\begin{enumerate}
    \item {\it Times-series level:} For a long time-series, we can split the original time-series into shorter time-series examples, each of length $T$, then select $n$ of those shorter examples to form the context$\{\*y^{(i)}_{1:T}\}_{i=1}^n$ for the original time-series.
    \item {\it Dataset level:} For each dataset, we can group any $n$ segments of length $T$ from any of the time-series in that dataset, to form a context. For instance, a set of $n$ segments from any of the time-series from the Electricity dataset could be grouped to form a context $\{\*y^{(i)}_{1:T}\}_{i=1}^n$.
\end{enumerate}
Both time-series level and dataset level groupings guarantee that the grouped examples have similar patterns to borrow from each other.

{\bf Dataset Mixture.} We choose all datasets in \cref{tab:train_data} other than the four Wiki datasets to generate in-context examples for continued training. The Wiki datasets contain millions of time-series that correspond to a wide variety of articles, which need not be related to each other. In fact the Wiki dataset can be potentially clustered into groups of related articles, and the time-series for each group could be deemed to form a separate dataset. But we leave such preprocessing of the Wiki dataset for future work and leave these datasets out of our continued pretraining. 

For the remaining datasets, we set the number of examples in each context as $n=50$ and generate contexts from both time-series level and dataset level grouping. Note that if all the time-series in a dataset have a total of $N$ examples, then generating all ${N \choose n}$ such contexts is intractable. Therefore, we randomly generate $20N$ such groups of $n$ examples as our training contexts.

Following the original TimesFM paper, the training data loader samples 90\% real data and 10\% synthetic, with the real data mixture providing equal weights to the groups: hourly + sub-hourly, daily, weekly, and monthly datasets. Moreover, we provide equal weights to the two kinds of examples i.e., time-series level and dataset level.

\section{Experimental Results}
Following prior time-series foundation model papers like~\citep{das2023decoder, gruver2023large}, we compare the zero-shot performance of our proposal with that of supervised models, statistical models trained per dataset as well as other zero-shot models. Similar to prior works, we report our results on a subset of Monash datasets~\citep{godahewa2021monash} and the ETT datasets~\citep{zhou2021informer} that have not been seen by our model or the TimesFM (base) model. 

\begin{figure}[ht]
    \centering
    \begin{subfigure}[t]{0.48\textwidth}
        \centering
        \includegraphics[width=\textwidth]{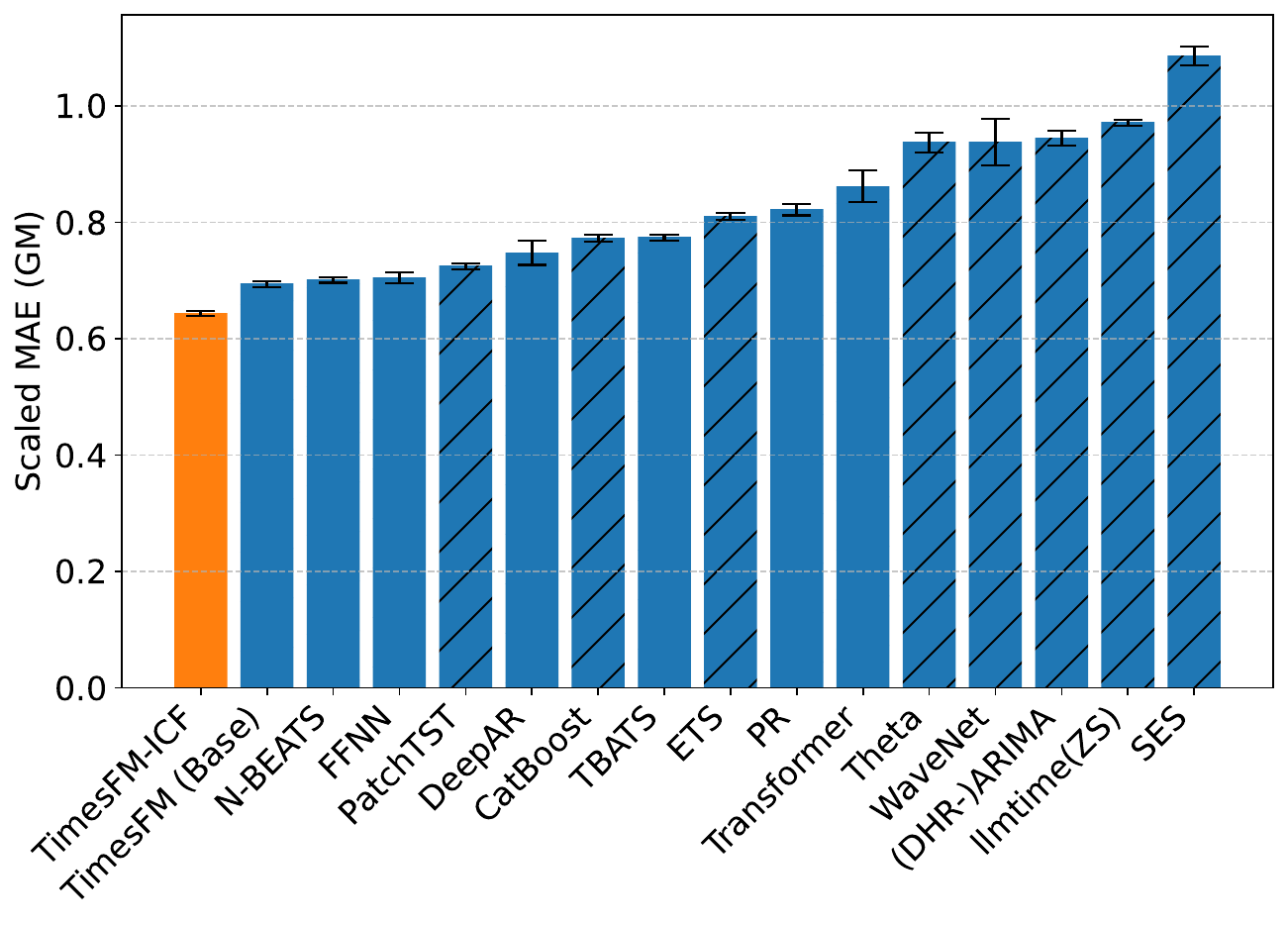}
        \caption{Monash}
        \label{fig:monashExperiments}
    \end{subfigure}
    \begin{subfigure}[t]{0.48\textwidth}
        \centering
        \includegraphics[width=\textwidth]{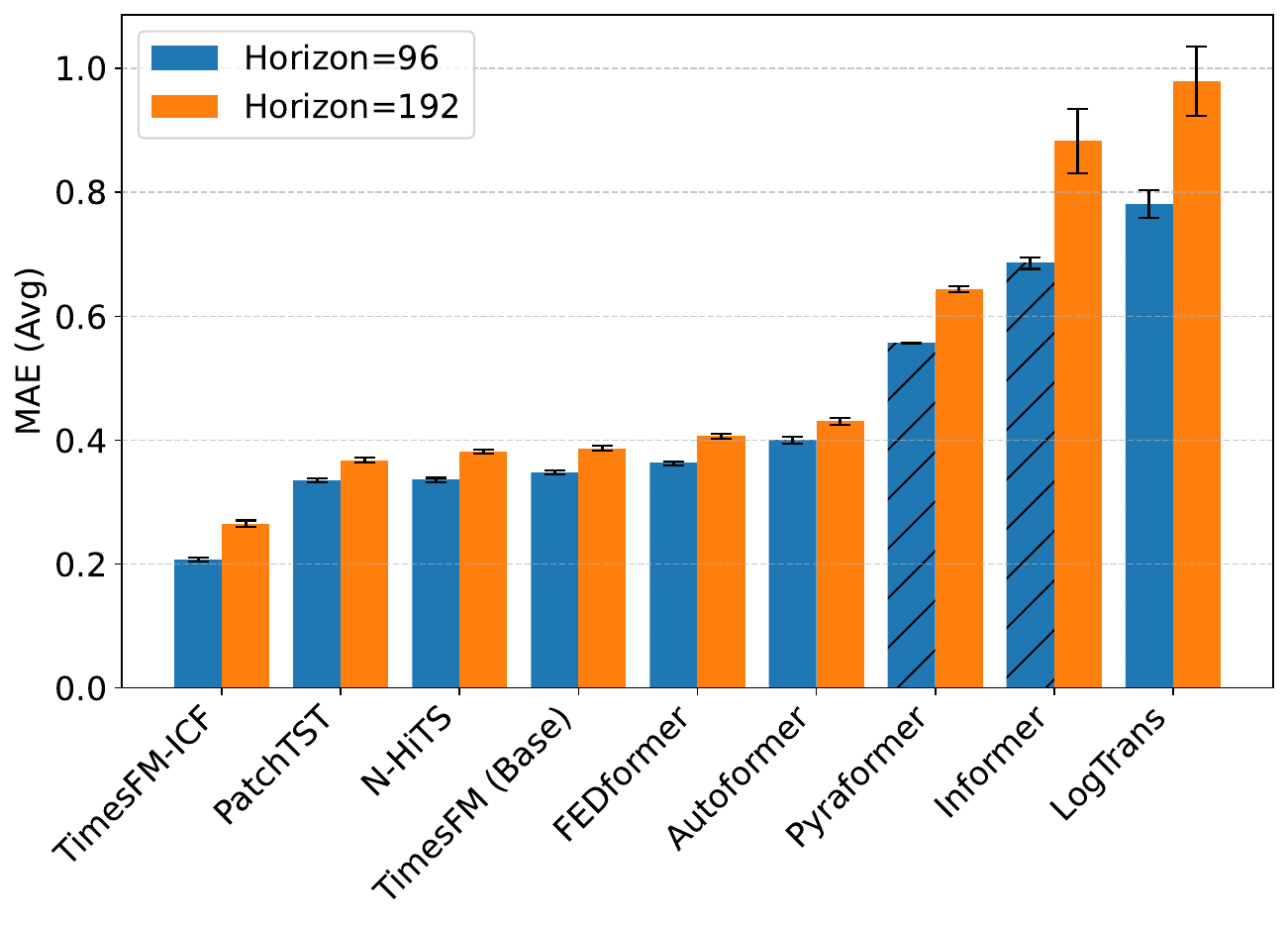}
        \caption{ETT}
        \label{fig:ett}
    \end{subfigure}
    \caption{In (a), we report the geometric mean of scaled MAE for Monash datasets. We include all official Monash baselines as well as \ours, \oursbase. \oursbase~yields a 7\% improvement over the next best baseline. We also report one standard error similar to~\citep{das2023decoder}. In (b), we report the average MAE numbers for 4 datasets ETTh1, ETTh2, ETTm1 and ETTm2. Similar to prior work like~\citep{nie2022time}, the numbers are reported for rolling validation over the test split which makes up the last 1/5th of time-points in each dataset. We also report one standard error. \ours~yields a marked improvement of at least 25\% over other baselines.}
\end{figure}

\subsection{Out-of-domain Forecasting on Monash}
\label{sec:monash_exp}
Monash archive~\citep{godahewa2021monash} is a collection of 30 datasets of different training and prediction lengths that covers granularities ranging from minutes to years and domains including finance, demand forecasting, weather and traffic. The archive reports four official metrics for several statistical baselines such as Exponential Smoothing(ETS) and ARIMA, as well as supervised ML baselines like CatBoost~\citep{prokhorenkova2018catboost}, DeepAR~\citep{salinas2020deepar} and WaveNet~\citep{oord2016wavenet}. We report our results on the 18 datasets that were also considered for zero-shot forecasting in~\citet{das2023decoder}. We provide more details in Appendix~\ref{app:monash}.

The datasets contain time-series with vastly different scales, so we cannot aggregate the raw metrics. Therefore, following prior works~\citep{gruver2023large, das2023decoder}, we calculate the MAE for all methods and normalize them by the MAE achieved by a naive baseline that just repeats the last time-point's value in the history for the whole horizon. Then we report the Geometric Mean of these scaled MAE values across all datasets. Note that when dealing with normalized metrics it is better to report the Geometric Mean~\citep{fleming1986not}. We borrow the official numbers for all baselines from~\citep{godahewa2021monash} except for \oursbase (we evaluate our base model) and LLMTime (we use the precomputed output from the original paper).

The results are summarized in Figure~\ref{fig:monashExperiments}. We can see that \ours~performs the best followed by \oursbase~and N-BEATS. It can be seen that \ours~yields a 7\% improvement over the closest supervised baseline (N-BEATS), which has been trained per dataset. More importantly, we obtain a 7\% improvement over \oursbase, thus showing the value of in-context fine-tuning for time-series foundation models. Note that \ours, \oursbase~and LLMTime are the only zero-shot methods in this benchmark.

\subsection{Out-of-domain Forecasting on ETT}
A group of long horizon datasets have been commonly used for benchmarking (mainly) transformer based deep learning algorithms starting from~\citep{zhou2021informer}. Some of the datasets in these benchmarks are in our pretraining datasets (like Electricity and Traffic). Therefore, for the interest of zero-shot evaluation we use the 4 Electricity Transformer Temperature (ETT) datasets, specifically ETTh1, ETTh2 (hourly) and ETTm1, ETTm2 (15 min). 

In terms of baselines, following \citep{das2023decoder}, we compare against Informer~\citep{zhou2021informer} and subsequent works like Pyraformer~\citep{liu2021pyraformer}, FEDFormer~\citep{zhou2022fedformer}, PatchTST~\citep{nie2022time}. We also compare with N-HiTS~\citep{challu2022nhits} which yields an improvement over N-BEATS~\citep{oreshkin2019n} for these datasets. Similar to ~\citet{das2023decoder}, we focus on the task of predicting horizon lengths 96, 192 given a history of 512 time-steps. We provide rolling validation numbers for the test time-period which consists the last 1/5-th of the time-points. This is standard for these benchmarks~\citep{nie2022time}, where the datasets are split into train:validation:test in the ratio 7:1:2.

We present the MAE obtained for horizon lengths 96 and 192 averaged over the 4 datasets in Figure~\ref{fig:ett}. Note that since the MAE is computed on scaled datasets in this benchmark~\citep{zhou2021informer}, we can directly report the arithmetic mean across datasets. We see that \ours~yields a marked improvement of more than 25\% on mean MAE over the nearest baseline. PatchTST, N-HiTS and \oursbase~perform similarly and are much better than the other baselines. In this case, all the datasets have in-context examples with enough time-points to cover $T$ time-steps, unlike in Monash where 9 out of 18 datasets have time-series of length less than 512 time-steps. Therefore, we can see more value from in-context fine-tuning. We provide a more fine-grained analysis with the number of in-context examples on ETTh datasets in Sections~\ref{sec:num_examples} and~\ref{sec:long_context}.

\begin{figure}[ht]
    \centering
    \begin{subfigure}[t]{0.45\textwidth}
        \centering
        \includegraphics[width=\textwidth]{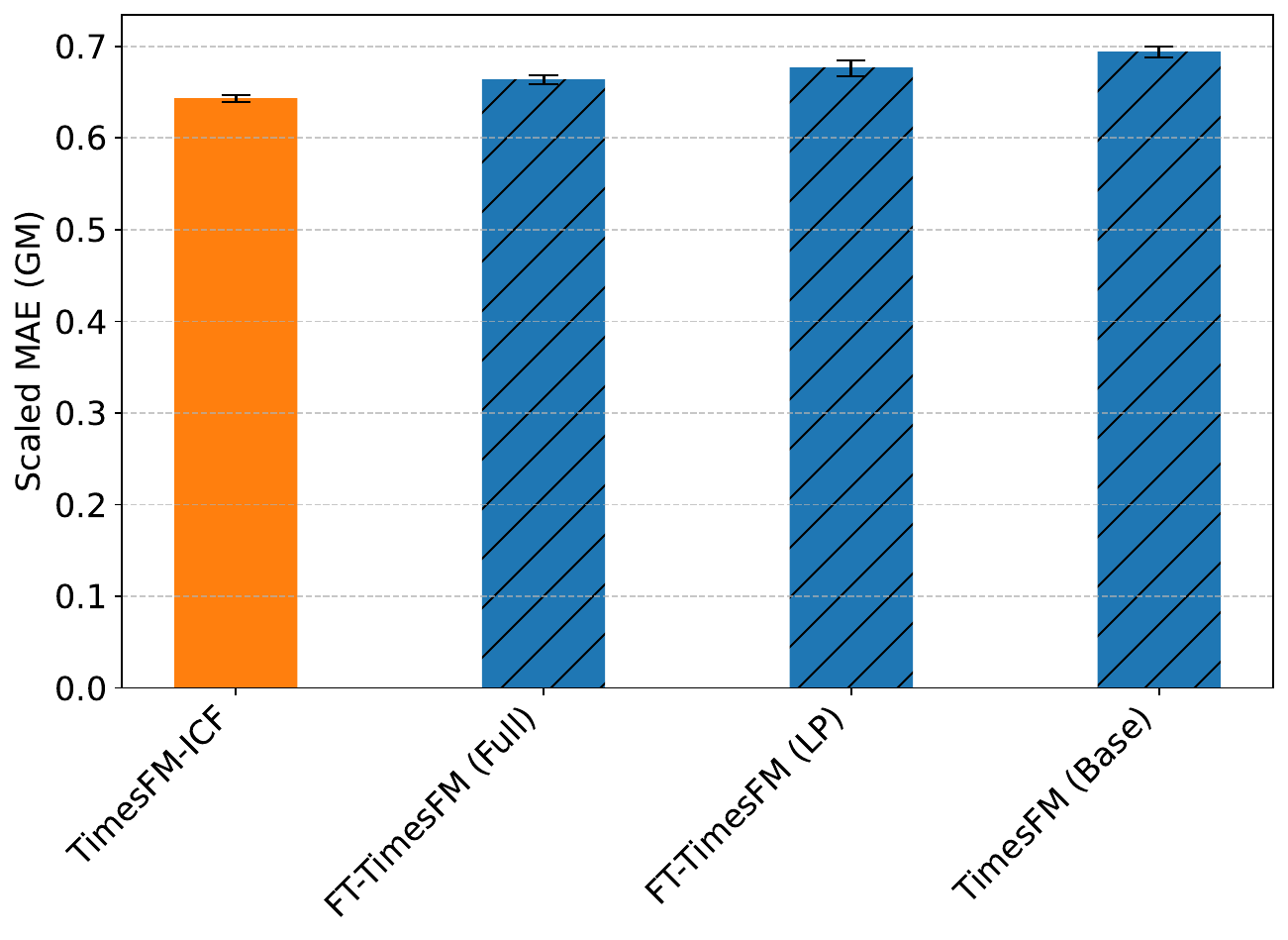}
        \caption{Fine-tuning per Dataset}
        \label{fig:monashPerDSFinetune}
    \end{subfigure}
    \begin{subfigure}[t]{0.45\textwidth}
        \centering
        \includegraphics[width=\textwidth]{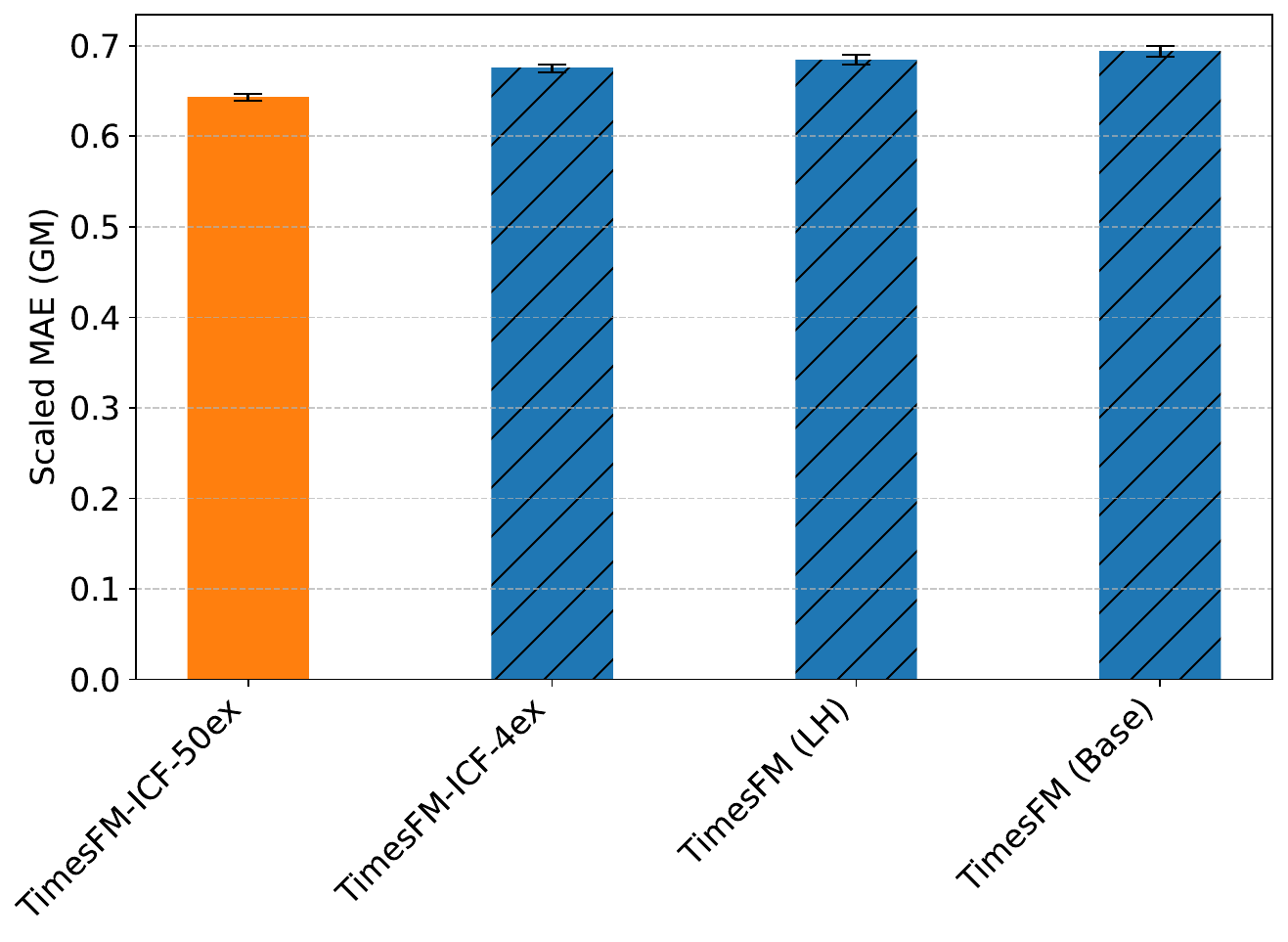}
        \caption{Longer History}
        \label{fig:monash-lc}
    \end{subfigure}
    \caption{In (a), we report the geometric mean of scaled MAE across the Monash datasets. FT-TimesFM corresponds to fine-tuning the original \oursbase~model per dataset either (1) Full fine-tune or (2) Linear Probed (see Section~\ref{sec:finetune}). We can see that \ours~is clearly better than FT-TimesFM models even though it is zero-shot. In (b), we compare \ours~with a base TimesFM model trained with a longer maximum supported history of 2048 time-points. We can see that \ours~ performs better than TimesFM (LH) in terms of the scaled MAE (GM) metric on Monash. This is further discussed in Section~\ref{sec:long_context}.}
\end{figure}

\subsection{Comparison with Fine-tuning per dataset}
\label{sec:finetune}
One of the main motivations of this work was to see whether we can recover the gains from fine-tuning foundation models on the target domain without doing any gradient updates. Therefore, in this section, we compare against a very strong baseline: for every dataset in our Monash benchmark from Section~\ref{sec:monash_exp} we fine-tune the \oursbase~model on the training set and evaluate it on the test set. We do two kinds of fine-tuning (1) we update all the model weights which we will refer to as FT-TimesFM (Full) (2) we hold all the transformer layer fixed while only the input and output residual blocks are fine-tuned, which we will refer to as FT-TimesFM (LP).\footnote{LP is meant to stand for Linear Probing even though here we are tuning the MLP layers.}

The aggregated scaled MAE numbers are presented in Figure~\ref{fig:monashPerDSFinetune}. \ours~actually yields close to 3\% improvement over FT-TimesFM (Full) which is already a 4\% improvement over \oursbase. This  shows that in-context fine-tuning can sometimes be better than per-dataset fine-tuning, even though we do not perform any gradient updates! The advantages of our method are further highlighted by the fact the total time required for fine-tuning on all datasets is \textit{115 minutes} (not including job scheduling times) for the cheaper FT-TimesFM (LP) method while the total inference time for \ours~is merely 4 minutes.\footnote{The inference numbers are reported on \href{https://cloud.google.com/tpu/docs/v5e-training}{TPUv5e} with 8 tensor cores.} 

While this is surprising, we believe that one reason could be that in many of the smaller datasets in Monash, fine-tuning the weights of a foundation model can actually lead to catastrophic forgetting of the learnt patterns which is also observed in LLMs~\citep{luo2023empirical}. Indeed on the smaller datasets like tourism yearly, bitcoin and us births, \ours~is better than FT-TimesFM and vice versa on larger datasets like Australian electricity demand. We provide per dataset metrics and more details about the fine-tuning in Section~\ref{app:ft}.

\subsection{Ablation}
We now present two important ablation studies that justify the benefits of in-context examples, as well as the advantages of our technique versus others like training longer-history models.

\subsubsection{Number of examples}
\label{sec:num_examples}
\begin{figure}[ht]
    \centering
    \includegraphics[width=.6\textwidth]{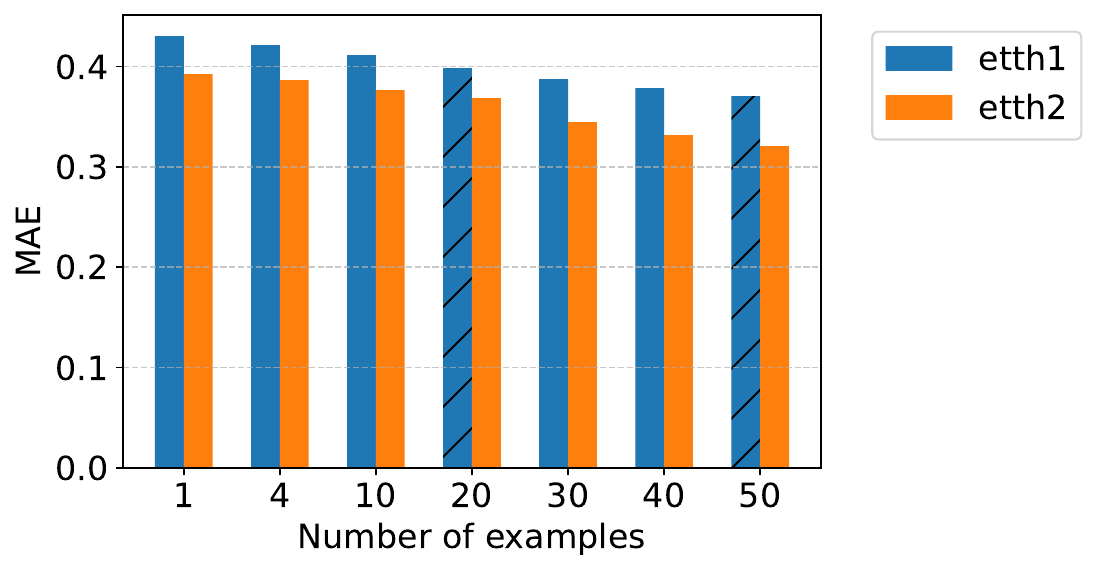}
    \caption{The performance of the model gets better with increasing number of in-context examples on ETTh1 and ETTh2.}
    \label{fig:ettVaryNumChunks}
\end{figure}

The number of in-context examples is an important consideration that dictates the performance of our model. We perform an ablation where we vary the number of in-context examples from 1 to the maximum during our training i.e. $n=50$. The corresponding results are reported on the ETTh test set in Figure~\ref{fig:ettVaryNumChunks}. We can see a monotonic increase in performance with more in-context examples. We chose to perform this ablation on the ETT datasets since, unlike the Monash datasets, all time-series are big enough to provide complete in-context examples of length $T$, which makes it easier to perform this experiment.

\subsubsection{Longer History}
\label{sec:long_context}

In this section, we compare the performance of \ours~with a version of \oursbase~trained with a longer history $L=2048$ which we will refer to as TimesFM (LH). We provide the aggregate scaled MAE on Monash datasets in Figure~\ref{fig:monash-lc} where we include two versions of \ours, one with 4 in-context examples (TimesFM-ICF-4ex) and one with 50 in-context examples (TimesFM-ICF-50ex). We can see that TimesFM (LH) yields a modest 1\% improvement over \oursbase~(which has a maximum history of 512) while TimesFM-ICF-50ex yields a 7\% improvement. Even TimesFM-ICF-4ex which uses the same total context length for all in-context examples as TimesFM (LH) is 3\% better than the baseline. 

This shows that our technique of in-context fine-tuning can be more effective than training a longer history model, especially when there is a mix of short-history and long-history time-series. This is because, for in-context fine-tuning, many short time-series can be packed as in-context examples inside the context, while for the case of usual long history training such time-series will just be padded and most of the context is wasted. As shown in the detailed results in Appendix~\ref{app:more_exp}, the long history model performs better on longer datasets like australian electricity demand, but degrades on shorter datasets like cif and tourism yearly.

\section{Conclusion}

In this paper, we introduce and study a methodology for in-context fine-tuning of a time-series foundation model for forecasting. In particular, we start with a base foundation model and adapt it to be able to effectively utilize, at inference time, not just the history of the target time-series for forecasting, but also in-context examples from related time-series.
Our results show that in-context fine-tuning can lead
to significantly better zero-shot performance on popular forecasting benchmarks compared to the base foundation model and state-of-the-art  supervised models. Furthermore, it even outperforms a version of the base foundation model that is explicitly fine-tuned on the target domain.

While we have chosen a specific base time-series foundation model (TimesFM) for our in-context fine-tuning approach, it would be an interesting direction of future work to study these adaptations for other base foundation models. It would also be interesting to study better forms of relative positional encodings specifically designed for handling in-context examples and length generalization.

\newpage

\printbibliography

\newpage

\appendix
\section{Appendix}

\subsection{More Details about our Model and Baselines}
\label{app:model}

{\bf Monash Baselines.} For the results on Monash datasets, we borrow the official numbers from~\citep{godahewa2021monash}. For LLMTime~\citep{gruver2023large} we use the pre-computed outputs supplied by the original authors. 

We also add the PatchTST~\citep{nie2022time} as a baseline for this benchmark because it is the best performing baseline (only worse than our models) in the ETT datasets. For this model we use the hyperparameters used by original paper for the ETTh datasets~\footnote{\url{https://github.com/yuqinie98/PatchTST/blob/main/PatchTST_supervised/scripts/PatchTST/etth1.sh}}.

{\bf ETT Baselines.} On the ETT datasets, the baseline numbers (except TimesFM (base)) are borrowed from the official numbers reported in Table 2 of~\citep{das2023longterm}. We evaluate the base model, TimesFM (base) as well as our method in a rolling validation manner on the test splits to obtain the corresponding metrics.

{\bf TimesFM (base).} Following~\citet{das2023decoder}, we train a 200M model with 16 attention heads, 20 layers, a input patch length of 32 and output patch length of 128. The model dimension is set to 1280. We use the learning rate schedule in~\citep{vaswani2017attention} with peak learning rate of $5e-4$. The hidden dims of both the residual block and the FFN in the transformer layers are set as the same as model dimensions. We keep layer norm in transformer layers but not in the residual blocks. The only difference between the model in~\citet{das2023decoder} and our base model is that we use NoPE instead of teh absolute positional encoding~\citep{vaswani2017attention}. As we have mentioned before, this leads to no loss in accuracy while being easier to extend to our in-context fine-tuning setting.

{\bf Fine-tuning Per Dataset.} On the Monash benchmark, we also compare with TimesFM (base) fine-tuned on the train set for every dataset and the forecasting on the corresponding test set. For all our fine-tuning runs, we use a batch size of 16 and a maximum of 10k iterations. Note that this means that the fine-tuned model will see many more training examples than the in-context examples given to our model. For the fine-tuning runs, we use the same decoder only loss function that was used in the original pretraining of TimesFM (base), the only difference is that the training is not restricted to the training set of one dataset. We do two kinds of fine tuning:

\begin{itemize}
    \item \textit{Full:} All weights in the model are updated during fine-tuning.
    \item \textit{Linear Probing (LP):} We hold the transformer weights fixed and only update the parameters in the input and output residual blocks.
\end{itemize}

{\bf \ours.} We continue to train \ours~model from \oursbase. Therefore, most of the parameters in the model remain the same. Here, are the key training details that are unique to \ours:

\begin{itemize}
    \item \textit{Separator Token:} We have a trainable separator token that is also updated during the continued pretraining. The token is nothing but a learnt embedding whose dimension is equal to the model dimension i.e. 1280 in our case.
    \item \textit{Number of Examples:} We use a maximum of $n=50$ in-context examples for each context during training.
    \item \textit{Padding:} In short datasets like M4 yearly and quarterly, each time-series might have number of time-points much less than $T = 640$. Sometimes the number of time-points are even less than our input patch length $p=32$. For such cases, a whole time-series can fit into one of the $n$ examples and they are preprocessed in the following manner:
        \begin{itemize}
            \item If the length of the time-series $l$ is less than $p$, we left pad with $k$ padding time-points such that $p < k + l < 2p$. This is because we want the decoder only model to predict something meaningful for the second patch after seeing the first patch and if not, is penalized by the loss on the second patch. If the $l > p$, we do not need to perform this left padding.
            \item Lastly, we right pad such that the length of the total padded example is $T=640$.
            \item Note that the last patch in such examples would be padded from the right, i.e., they will have real time-series values for the first few points and padding for the rest. We make sure that such incomplete from the right patches are not attended by subsequent tokens belonging to examples coming after.
        \end{itemize}
\end{itemize}

The pretraining datasets are detailed in Table~\ref{tab:train_data}.

\begin{table}[!ht]
    \centering
    \caption{List of datasets included in pretraining. All datasets except the Wiki datasets are also repurposed for continued pretraining with in-context examples.}
    \resizebox{0.7\textwidth}{!}{
    \begin{tabular}{llrr}
    \toprule
    \multicolumn{1}{c}{Dataset} & \multicolumn{1}{c}{Granularity} & \multicolumn{1}{c}{\# Time series} & \multicolumn{1}{c}{\# Time points} \\ \midrule
    Synthetic & & 3,000,000 & 6,144,000,000  \\
    Electricity & Hourly & 321 & 8,443,584  \\
    Traffic & Hourly & 862 & 15,122,928  \\
    Weather~\citep{zhou2021informer} & 10 Min & 42 & 2,213,232  \\
    Favorita Sales & Daily & 111,840 & 139,179,538  \\
    LibCity~\citep{libcitylong} & 15 Min & 6,159 & 34,253,622  \\
    M4 hourly & Hourly & 414 & 353,500  \\
    M4 daily & Daily & 4,227  & 9,964,658  \\
    M4 monthly & Monthly & 48,000 & 10,382,411  \\
    M4 quarterly & Quarterly & 24,000 & 2,214,108  \\
    M4 yearly & Yearly & 22,739 & 840,644  \\
    Wiki hourly & Hourly & 5,608,693 & 239,110,787,496  \\
    Wiki daily & Daily & 68,448,204 & 115,143,501,240  \\
    Wiki weekly & Weekly & 66,579,850 & 16,414,251,948  \\
    Wiki monthly & Monthly & 63,151,306 & 3,789,760,907  \\
    Trends hourly & Hourly & 22,435 & 393,043,680  \\
    Trends daily & Daily & 22,435 & 122,921,365  \\
    Trends weekly & Weekly & 22,435 & 16,585,438  \\
    Trends monthly & Monthly & 22,435 & 3,821,760 \\
    \bottomrule
    \end{tabular}}
    \label{tab:train_data}
\end{table}

\subsection{Detailed Metrics on Monash and ETT}
\label{app:more_exp}

\subsubsection{Monash}
\label{app:monash}
Table~\ref{table:monash-raw} presents the per-dataset MAE numbers of \ours~ against other supervised and zero-shot methods on Monash.

\begin{table}[ht]
\centering
\caption{MAE of \ours~ against other supervised and zero-shot methods on Monash.}
\label{table:monash-raw}
\resizebox{\linewidth}{!}{%
\begin{tabular}{llllllllllllllllll}
\toprule
 & (DHR-)ARIMA & CatBoost & DeepAR & ETS & FFNN & N-BEATS & Naive & PR & PatchTST & SES & TBATS & Theta & TimesFM (Base) & TimesFM-ICF & Transformer & WaveNet & llmtime(ZS) \\
\midrule
australian electricity demand & 1045.92 & 241.77 & 302.41 & 1282.99 & 258.76 & 213.83 & 659.60 & 247.18 & 248.35 & 659.60 & 370.74 & 665.04 & 426.12 & 338.98 & 231.45 & 227.50 & 459.96 \\
bitcoin & 3.62e+18 & 1.93e+18 & 1.95e+18 & 1.10e+18 & 1.45e+18 & 1.06e+18 & 7.78e+17 & 6.66e+17 & 1.84e+18 & 5.33e+18 & 9.90e+17 & 5.33e+18 & 1.90e+18 & 9.58e+17 & 2.61e+18 & 2.46e+18 & 1.75e+18 \\
fred md & 2957.11 & 2475.68 & 4264.36 & 2041.42 & 2339.57 & 2557.80 & 2825.67 & 8921.94 & 2005.86 & 2798.22 & 1989.97 & 3492.84 & 2514.63 & 2021.52 & 4666.04 & 2508.40 & 2013.49 \\
nn5 daily & 4.41 & 4.22 & 3.94 & 3.72 & 4.06 & 4.92 & 8.26 & 5.47 & 5.56 & 6.63 & 3.70 & 3.80 & 3.57 & 3.74 & 4.16 & 3.97 & 9.39 \\
pedestrian counts & 635.16 & 43.41 & 44.78 & 216.50 & 46.41 & 66.84 & 170.88 & 44.18 & 45.90 & 170.87 & 222.38 & 170.94 & 42.55 & 43.71 & 47.29 & 46.46 & 70.20 \\
saugeenday & 22.38 & 21.28 & 23.51 & 30.69 & 22.98 & 27.92 & 21.50 & 25.24 & 21.52 & 21.50 & 22.26 & 21.49 & 30.54 & 24.91 & 28.06 & 22.17 & 28.63 \\
traffic hourly & 0.04 & 0.02 & 0.01 & 0.03 & 0.01 & 0.02 & 0.03 & 0.02 & 0.01 & 0.03 & 0.04 & 0.03 & 0.01 & 0.01 & 0.01 & 0.02 & 0.03 \\
us births & 526.33 & 441.70 & 424.93 & 419.73 & 557.87 & 422.00 & 1152.67 & 574.93 & 556.23 & 1192.20 & 399.00 & 586.93 & 446.49 & 399.74 & 452.87 & 504.40 & 459.43 \\
weather & 2.45 & 2.51 & 2.02 & 2.35 & 2.09 & 2.34 & 2.36 & 8.17 & 2.12 & 2.24 & 2.30 & 2.51 & 1.98 & 2.10 & 2.03 & 2.29 & 2.32 \\
cif 2016 & 469059.49 & 603551.30 & 3200418.00 & 642421.42 & 1495923.44 & 679034.80 & 386526.37 & 563205.57 & 271198.00 & 581875.97 & 855578.40 & 714818.58 & 438028.90 & 647255.33 & 4057973.04 & 5998224.62 & 715086.33 \\
covid deaths & 85.77 & 475.15 & 201.98 & 85.59 & 144.14 & 158.81 & 353.71 & 347.98 & 246.55 & 353.71 & 96.29 & 321.32 & 124.86 & 113.78 & 408.66 & 1049.48 & 304.68 \\
hospital & 19.60 & 19.17 & 18.25 & 17.97 & 22.86 & 20.18 & 24.07 & 19.24 & 18.52 & 21.76 & 17.43 & 18.54 & 17.95 & 17.26 & 36.19 & 19.35 & 24.62 \\
nn5 weekly & 15.38 & 15.29 & 14.69 & 15.70 & 15.02 & 14.19 & 16.71 & 14.94 & 15.38 & 15.66 & 14.98 & 15.30 & 14.15 & 15.38 & 20.34 & 19.34 & 15.91 \\
solar weekly & 839.88 & 1513.49 & 721.59 & 1131.01 & 1050.84 & 1172.64 & 1729.41 & 1044.98 & 1525.59 & 1202.39 & 908.65 & 1210.83 & 1380.09 & 1424.71 & 576.35 & 1996.89 & 2049.09 \\
tourism monthly & 2536.77 & 2537.04 & 1871.69 & 2004.51 & 2022.21 & 2003.02 & 5636.83 & 2187.28 & 2587.16 & 5302.10 & 2940.08 & 2069.96 & 3406.55 & 2018.07 & 2146.98 & 2095.13 & 4724.94 \\
tourism quarterly & 10475.47 & 10267.97 & 9511.37 & 8925.52 & 8981.04 & 8640.56 & 15845.10 & 9092.58 & 13271.98 & 15014.19 & 9972.42 & 7656.49 & 9535.86 & 8202.19 & 9521.67 & 9137.12 & 14121.09 \\
tourism yearly & 95033.24 & 79567.22 & 71471.29 & 94818.89 & 79593.22 & 70951.80 & 99456.05 & 82682.97 & 99574.68 & 95579.23 & 94121.08 & 90653.60 & 75955.39 & 80365.15 & 74316.52 & 69905.47 & 140081.78 \\
traffic weekly & 1.22 & 1.17 & 1.18 & 1.14 & 1.15 & 1.11 & 1.19 & 1.13 & 1.15 & 1.12 & 1.17 & 1.13 & 1.06 & 1.09 & 1.42 & 1.20 & 1.17 \\
\midrule
Scaled MAE (GM) & 0.945 & 0.773 & 0.748 & 0.810 & 0.704 & 0.700 & 1.000 & 0.822 & 0.724 & 1.086 & 0.774 & 0.937 & 0.694 & 0.643 & 0.862 & 0.938 & 0.971 \\

\bottomrule
\end{tabular}
}
\end{table}

\subsubsection{ETT}
\label{app:ett}

\begin{table}[ht]
\centering
\caption{MAE of \ours~against other baselines on ETT}
\label{table:ett}
\resizebox{\linewidth}{!}{%
\begin{tabular}{llrrrrrrrrr}
\toprule
 &  & Autoformer & FEDformer & Informer & LogTrans & N-HiTS & PatchTST & Pyraformer & TimesFM (Base) & TimesFM-ICF \\
\midrule
\multirow[t]{2}{*}{avg} & 96 & 0.400 & 0.362 & 0.686 & 0.781 & 0.336 & 0.335 & 0.556 & 0.348 & 0.207 \\
 & 192 & 0.430 & 0.406 & 0.883 & 0.979 & 0.381 & 0.368 & 0.643 & 0.387 & 0.265 \\
\cline{1-11}
\multirow[t]{2}{*}{etth1} & 96 & 0.446 & 0.415 & 0.769 & 0.740 & 0.393 & 0.401 & 0.612 & 0.398 & 0.263 \\
 & 192 & 0.457 & 0.446 & 0.786 & 0.824 & 0.436 & 0.429 & 0.681 & 0.427 & 0.330 \\
\cline{1-11}
\multirow[t]{2}{*}{etth2} & 96 & 0.368 & 0.374 & 0.952 & 1.197 & 0.345 & 0.337 & 0.597 & 0.350 & 0.206 \\
 & 192 & 0.434 & 0.446 & 1.542 & 1.635 & 0.401 & 0.376 & 0.683 & 0.392 & 0.265 \\
\cline{1-11}
\multirow[t]{2}{*}{ettm1} & 96 & 0.492 & 0.390 & 0.560 & 0.546 & 0.350 & 0.346 & 0.510 & 0.369 & 0.207 \\
 & 192 & 0.495 & 0.415 & 0.619 & 0.700 & 0.383 & 0.370 & 0.537 & 0.405 & 0.265 \\
\cline{1-11}
\multirow[t]{2}{*}{ettm2} & 96 & 0.293 & 0.271 & 0.462 & 0.642 & 0.255 & 0.256 & 0.507 & 0.274 & 0.152 \\
 & 192 & 0.336 & 0.318 & 0.586 & 0.757 & 0.305 & 0.296 & 0.673 & 0.323 & 0.201 \\
\cline{1-11}
\bottomrule
\end{tabular}
}
\end{table}

Table~\ref{table:ett} presents the MAE numbers of \ours~against other methods on ETTh1, ETTh2, ETTm1 and ETTm2 respectively, with forecasting horizons of 96 and 192 respectively.

\subsection{varying the number of in-context examples}
Table~\ref{table:ett-vary-num-chunks} and~\ref{table:monash-vary-num-chunks} shows the accuracy metric numbers of \ours~on ETT and Monash respectively when different numbers of in-context examples are used.

\begin{table}[ht]
\centering
\caption{MAE of \ours~on ETT with different numbers of in-context examples.}
\label{table:ett-vary-num-chunks}
\begin{tabular}{lrrrrrrr}
\toprule
Number of in-context examples & 1 & 4 & 10 & 20 & 30 & 40 & 50 \\
\midrule
etth1 & 0.430 & 0.421 & 0.411 & 0.398 & 0.387 & 0.378 & 0.371 \\
etth2 & 0.392 & 0.386 & 0.377 & 0.368 & 0.344 & 0.331 & 0.320 \\
Average MAE & 0.411 & 0.404 & 0.394 & 0.383 & 0.366 & 0.354 & 0.345 \\
\bottomrule
\end{tabular}
\end{table}

\begin{table}[ht]
\centering
\caption{Scaled MAE (GM) of \ours~on Monash with different numbers of in-context examples.}
\label{table:monash-vary-num-chunks}
\begin{tabular}{lrrrrrrrr}
\toprule
 Number of in-context examples & 1 & 4 & 5 & 10 & 20 & 30 & 40 & 50 \\
\midrule
Scaled MAE (GM) & 0.667 & 0.675 & 0.667 & 0.658 & 0.651 & 0.657 & 0.653 & 0.643 \\
\bottomrule
\end{tabular}
\end{table}

\subsection{Long History}
Table~\ref{table:monash-lc-normalized} and~\ref{table:monash-lc-raw} show respectively the aggregated (geometric mean of scaled MAE) and the raw MAE numbers on Monash of different TimesFM models, with the focus on the comparison between \ours~ and TimesFM (LH) which is a long-2048-history TimesFM model. We compare \ours in two different modes: (i) 50ex, in which the model has access to 50 in-context examples, and (ii) 4ex, in which the model has access to only 4 in-context examples. In mode (ii), the aggregate length of all in-context examples is the same as the length of the history used by TimesFM (LH). 

\begin{table}[ht]
\centering
\caption{Scaled MAE (GM) on Monash for long history length}
\label{table:monash-lc-normalized}
\begin{tabular}{lr}
\toprule
 & Scaled MAE (GM) \\
\midrule
TimesFM-ICF-50ex & 0.643 \\
TimesFM-ICF-4ex & 0.675 \\
TimesFM (LH) & 0.685 \\
TimesFM (Base) & 0.694 \\
\bottomrule
\end{tabular}
\end{table}

\begin{table}[ht]
\centering
\caption{Detailed breakdown of MAE on Monash for long history length}
\label{table:monash-lc-raw}
\resizebox{\linewidth}{!}{%
\begin{tabular}{llllll}
\toprule
 & TimesFM (LH) & TimesFM-ICF-4ex & TimesFM-ICF-50ex & TimesFM (Base) & naive \\
\midrule
australian electricity demand & 468.81 & 492.56 & 338.98 & 426.12 & 659.60 \\
bitcoin & 1.50e+18 & 1.32e+18 & 9.58e+17 & 1.90e+18 & 7.78e+17 \\
cif 2016 & 709069.14 & 477038.11 & 647255.33 & 438028.90 & 386526.37 \\
covid deaths & 151.64 & 131.75 & 113.78 & 124.86 & 353.71 \\
fred md & 1519.00 & 1795.34 & 2021.52 & 2514.63 & 2825.67 \\
hospital & 17.64 & 17.23 & 17.26 & 17.95 & 24.07 \\
nn5 daily & 3.52 & 3.74 & 3.74 & 3.57 & 8.26 \\
nn5 weekly & 15.05 & 14.80 & 15.38 & 14.15 & 16.71 \\
pedestrian counts & 43.96 & 46.30 & 43.71 & 42.55 & 170.88 \\
saugeenday & 25.87 & 29.40 & 24.91 & 30.54 & 21.50 \\
solar weekly & 1211.10 & 1324.05 & 1424.71 & 1380.09 & 1729.41 \\
tourism monthly & 2629.16 & 2155.61 & 2018.07 & 3406.55 & 5636.83 \\
tourism quarterly & 8595.55 & 8952.65 & 8202.19 & 9535.86 & 15845.10 \\
tourism yearly & 89423.79 & 85239.54 & 80365.15 & 75955.39 & 99456.05 \\
traffic hourly & 0.01 & 0.01 & 0.01 & 0.01 & 0.03 \\
traffic weekly & 1.08 & 1.09 & 1.09 & 1.06 & 1.19 \\
us births & 473.87 & 447.00 & 399.74 & 446.49 & 1152.67 \\
weather & 1.87 & 2.12 & 2.10 & 1.98 & 2.36 \\
\midrule
Scaled MAE (GM) & 0.685 & 0.675 & 0.643 & 0.694 & 1.000\\
\bottomrule
\end{tabular}
}
\end{table}

\subsection{Fine-tuning per Dataset}
\label{app:ft}

Table~\ref{table:monash-perds-gmeans}, \ref{table:monash-perds-raw} and~\ref{table:monash-perds-time} present the detailed accuracy and timing metrics to compare \ours~and FT-TimesFM on Monash. While \ours~is more accurate, it is also significantly faster than straighforward fine-tuning on the target dataset. Both are results of the \ours's in-context learning capability.

\begin{table}[ht]
\centering
\caption{Monash Per-Dataset Fine-tune (scaled MAE)}
\label{table:monash-perds-gmeans}
\begin{tabular}{lr}
\toprule
 & scaled MAE (GM) \\
\midrule
FT-TimesFM (Full) & 0.663 \\
FT-TimesFM (LP) & 0.676 \\
TimesFM-ICF & 0.643 \\
TimesFM (Base) & 0.694 \\
\bottomrule
\end{tabular}
\end{table}

\begin{table}[ht]
\centering
\caption{MAE on Monash of \ours~compared to models fine-tuned and evaluated on (the training and test set, respectively, within) each individual dataset within Monash}
\label{table:monash-perds-raw}
\resizebox{\linewidth}{!}{%
\begin{tabular}{llllll}
\toprule
 & FT-TimesFM (Full) & FT-TimesFM (LP) & TimesFM-ICF & TimesFM (Base) & naive \\
\midrule
australian electricity demand & 178.07 & 262.83 & 338.98 & 426.12 & 659.60 \\
bitcoin & 1.33e+18 & 1.43e+18 & 9.58e+17 & 1.90e+18 & 7.78e+17 \\
cif 2016 & 724237.52 & 1344910.30 & 647255.33 & 438028.90 & 386526.37 \\
covid deaths & 181.89 & 85.12 & 113.78 & 124.86 & 353.71 \\
fred md & 2296.35 & 2330.96 & 2021.52 & 2514.63 & 2825.67 \\
hospital & 19.53 & 18.86 & 17.26 & 17.95 & 24.07 \\
nn5 daily & 3.42 & 3.37 & 3.74 & 3.57 & 8.26 \\
nn5 weekly & 15.24 & 15.02 & 15.38 & 14.15 & 16.71 \\
pedestrian counts & 41.80 & 40.88 & 43.71 & 42.55 & 170.88 \\
saugeenday & 22.07 & 25.22 & 24.91 & 30.54 & 21.50 \\
solar weekly & 882.09 & 1610.53 & 1424.71 & 1380.09 & 1729.41 \\
tourism monthly & 2469.08 & 2069.82 & 2018.07 & 3406.55 & 5636.83 \\
tourism quarterly & 10140.35 & 10725.62 & 8202.19 & 9535.86 & 15845.10 \\
tourism yearly & 88210.94 & 85915.69 & 80365.15 & 75955.39 & 99456.05 \\
traffic hourly & 0.02 & 0.01 & 0.01 & 0.01 & 0.03 \\
traffic weekly & 1.19 & 1.12 & 1.09 & 1.06 & 1.19 \\
us births & 405.81 & 397.24 & 399.74 & 446.49 & 1152.67 \\
weather & 1.81 & 1.84 & 2.10 & 1.98 & 2.36 \\
\midrule
Scaled MAE (GM) & 0.663 & 0.676 & 0.643 & 0.694 & 1.000 \\
\bottomrule
\end{tabular}
}
\end{table}

\begin{table}[ht]
\centering
\caption{Timing breakdown (in minutes) of forecasting \ours~ compared to individually fine-tuning then evaluating models on a per-dataset basis in Monash}
\label{table:monash-perds-time}
\begin{tabular}{lrrr}
\toprule
 & FT-TimesFM (Full) & FT-TimesFM (LP) & TimesFM-ICF \\
\midrule
australian electricity demand & 6.350 & 2.370 & 0.048 \\
bitcoin & 9.600 & 4.620 & 0.053 \\
cif 2016 & 8.610 & 4.230 & 0.069 \\
covid deaths & 26.470 & 9.520 & 0.178 \\
fred md & 10.310 & 6.020 & 0.077 \\
hospital & 15.720 & 3.610 & 0.347 \\
nn5 daily & 11.120 & 5.360 & 0.076 \\
nn5 weekly & 9.220 & 3.950 & 0.081 \\
pedestrian counts & 17.120 & 12.050 & 0.063 \\
saugeenday & 9.440 & 4.090 & 0.048 \\
solar weekly & 9.040 & 5.030 & 0.085 \\
tourism monthly & 6.780 & 4.120 & 0.209 \\
tourism quarterly & 11.200 & 6.140 & 0.226 \\
tourism yearly & 10.350 & 5.160 & 0.288 \\
traffic hourly & 20.250 & 16.920 & 0.413 \\
traffic weekly & 7.700 & 11.790 & 0.428 \\
us births & 10.190 & 5.580 & 0.047 \\
weather & 7.540 & 4.910 & 1.394 \\
\midrule
Total & 207.010 & 115.470 & 4.130\\
\bottomrule
\end{tabular}
\end{table}

\subsection{Illustrative Examples}
\label{app:illu}

\begin{figure}
    \centering
    \begin{subfigure}{0.9\textwidth}
    \includegraphics[width=\linewidth]{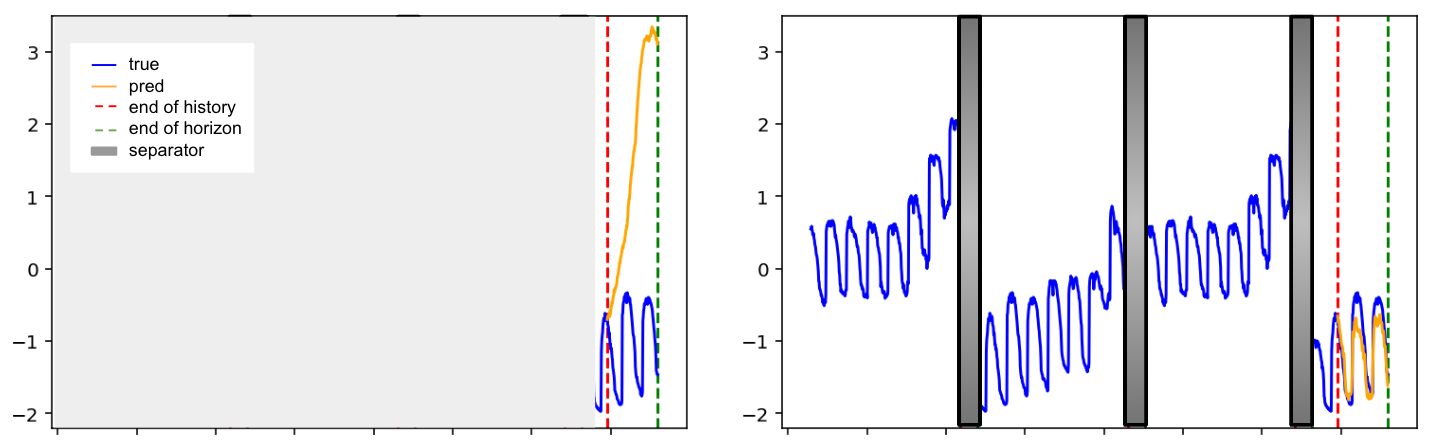}
    \caption{In-context examples help the history disambiguate between an increasing trend and an oscillating seasonality.}
    \end{subfigure}
    
    \begin{subfigure}{0.9\textwidth}
    \includegraphics[width=\linewidth]{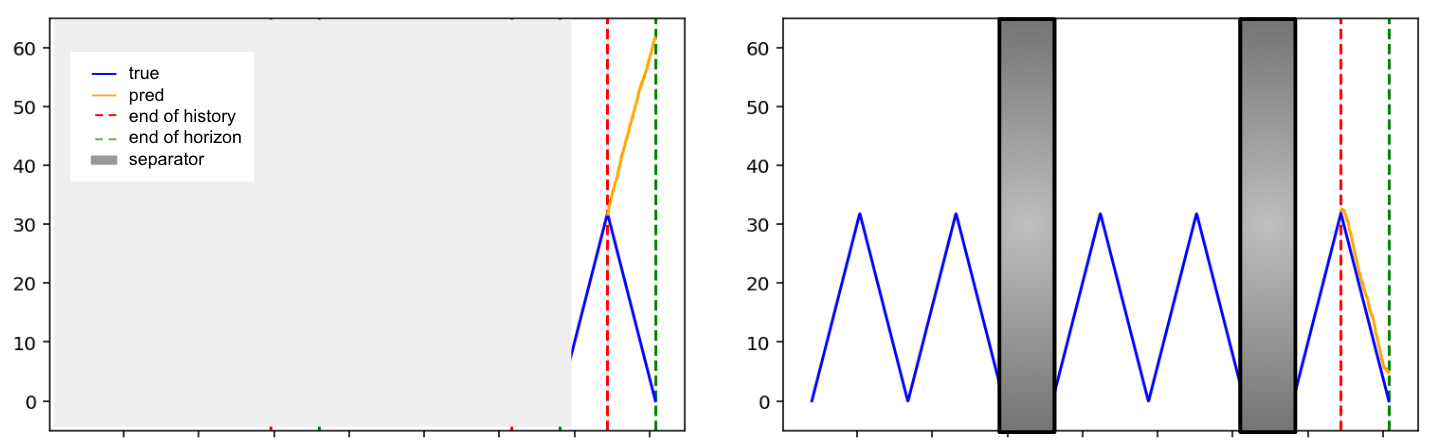}
    \caption{In-context examples help the history disambiguate between an increasing linear trend and a triangular wave.}
    \end{subfigure}
  \caption{Two illustrative examples on how in-context examples can help disambiguate the prediction tasks, that likely patterns based solely on the history can get proved or disproved by the patterns from the in-context examples.}
  \label{fig:app-illustrative-examples}
\end{figure}

We illustrate visually in Figure~\ref{fig:app-illustrative-examples} how in-context examples can help disambiguate the prediction tasks, by plotting the actual forecasts from \ours~with and without the in-context examples.
In the left two figures, the history is not sufficiently informative for the model to make an accurate prediction. By providing in-context examples together with this short history (see the right two figures), however, the model is able to make a more accurate forecast.

\end{document}

%% file: math_commands.tex
\usepackage{amsmath,amsfonts,bm}

\def\eqref#1{equation~\ref{#1}}

\def\ceil#1{\lceil #1 \rceil}

\def\1{\bm{1}}

\def\rb{{\textnormal{b}}}

\def\vmu{{\bm{\mu}}}
\def\vtheta{{\bm{\theta}}}

\def\vb{{\bm{b}}}

\DeclareMathAlphabet{\mathsfit}{\encodingdefault}{\sfdefault}{m}{sl}
\SetMathAlphabet{\mathsfit}{bold}{\encodingdefault}{\sfdefault}{bx}{n}

\def\sA{{\mathbb{A}}}
\def\sB{{\mathbb{B}}}
\def\sC{{\mathbb{C}}}
\def\sD{{\mathbb{D}}}
\def\sF{{\mathbb{F}}}
\def\sG{{\mathbb{G}}}
\def\sH{{\mathbb{H}}}
\def\sI{{\mathbb{I}}}
\def\sJ{{\mathbb{J}}}
\def\sK{{\mathbb{K}}}
\def\sL{{\mathbb{L}}}
\def\sM{{\mathbb{M}}}
\def\sN{{\mathbb{N}}}
\def\sO{{\mathbb{O}}}
\def\sP{{\mathbb{P}}}
\def\sQ{{\mathbb{Q}}}
\def\sR{{\mathbb{R}}}
\def\sS{{\mathbb{S}}}
\def\sT{{\mathbb{T}}}
\def\sU{{\mathbb{U}}}
\def\sV{{\mathbb{V}}}
\def\sW{{\mathbb{W}}}
\def\sX{{\mathbb{X}}}
\def\sY{{\mathbb{Y}}}
\def\sZ{{\mathbb{Z}}}

\DeclareMathOperator{\sign}{sign}

\let\ab\allowbreak